%% file: main.tex
\documentclass[sigconf]{acmart}
\renewcommand\footnotetextcopyrightpermission[1]{} % removes footnote with conference information in first column
\AtBeginDocument{%
  \providecommand\BibTeX{{%
    \normalfont B\kern-0.5em{\scshape i\kern-0.25em b}\kern-0.8em\TeX}}}

\usepackage{algorithmic}
\usepackage{algorithm}
\usepackage{balance}
\usepackage{bbding}
\usepackage{color}
\usepackage{bbm}
\definecolor{quill}{rgb}{0.69,0.61,0.85}
\definecolor{contentchange}{rgb}{0.9,0.61,0.5}

\input{symbols}
\begin{document}

\title{Style-A-Video: Agile Diffusion for Arbitrary Text-based Video Style Transfer}
\author{Nisha Huang$^{1,2}$\quad Yuxin Zhang$^{1,2}$\quad Weiming Dong$^{1,2}$\\ 
$^1$School of Artificial Intelligence, UCAS \quad$^2$MAIS, Institute of Automation, CAS\\ 
\url{https://github.com/haha-lisa/Style-A-Video}
}



\input{Figures/teaser}
\input{Sections/0_abstract}
\maketitle

\input{Sections/1_intro}

\input{Sections/2_related}

\input{Sections/3_method}

\input{Sections/4_experiments}
\input{Sections/6_conclusion}

% \begin{acks}
% This work was supported by National Key R\&D Program of China under No. 2020AAA0106200, by National Natural Science Foundation of China under Nos. 61832016, U20B2070, 6210070958, 62102162, and by Open Projects Program of NLPR.
% \end{acks}

\bibliographystyle{ACM-Reference-Format}
\balance
\bibliography{references}

\end{document}

%% file: symbols.tex
\newcommand{\shortname}{Style-A-Video}
\newcommand{\etal}{{\sl et al.}}

\newcommand{\textprompt}{\mathcal{T}}
\newcommand{\loss}{\mathcal{L}}

%% file: Figures/teaser.tex
\begin{teaserfigure}
    \centering
    \includegraphics[width=\linewidth]{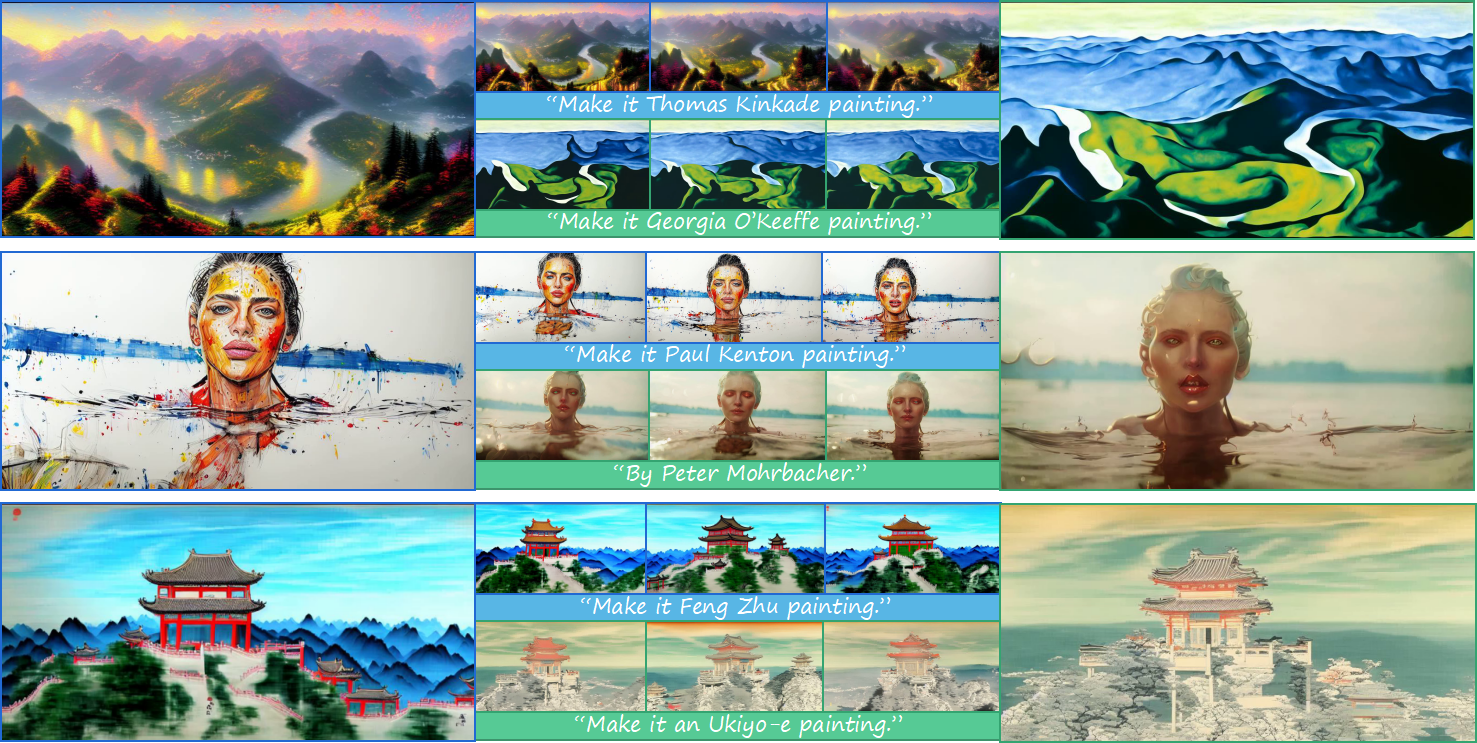}
    % \vspace{-0.2cm} 
    \caption{We propose a text-driven video stylization method \textbf{Style-A-Video} based on diffusion models.}
    % \vspace{-0.4cm} 
    \label{fig:teaser}
\end{teaserfigure}

%% file: Sections/0_abstract.tex
\begin{abstract}
Large-scale text-to-video diffusion models have demonstrated an exceptional ability to synthesize diverse videos. 
% However, applying these models directly to video stylization remains a challenge, 
% First, large-scale text-to-video datasets as well as training computational resources are difficult to obtain. 
% Second, given that the noise addition process on the input content is random and destructive, it is challenging to fulfill the stylization task's content preservation criteria.
% e.g. the lack of large-scale text-to-video datasets as well as the computational resources for training. 
However, due to the lack of extensive text-to-video datasets and the necessary computational resources for training, directly applying these models for video stylization remains difficult. 
Also, given that the noise addition process on the input content is random and destructive, fulfilling the style transfer task's content preservation criteria is challenging.
This paper proposes a zero-shot video stylization method named Style-A-Video, which utilizes a generative pre-trained transformer with an image latent diffusion model to achieve a concise text-controlled video stylization.
We improve the guidance condition in the denoising process, establishing a balance between artistic expression and structure preservation. 
Furthermore, to decrease inter-frame flicker and avoid the formation of additional artifacts, we employ a sampling optimization and a temporal consistency module. 
Extensive experiments show that we can attain superior content preservation and stylistic performance while incurring less consumption than previous solutions.
\end{abstract}

%% file: Sections/1_intro.tex
\section{INTRODUCTION}
\label{sec:intro}
% 1
% The manipulation of real-world images~\cite{cast,deng2021stytr2, Huang2022MGAD} and videos~\cite{mccnet_video, mccnet_v2_video} using computational approaches has advanced considerably.
% Style transfer tasks have gained an abundance of consideration and implementation among them.
% Rather than style images, we employ plainer and simply accessible text as style conditions in our work. As a result, real-world video style manipulation appears more versatile and creative.

% \IEEEPARstart{C}{omputational} methods for manipulating natural images~\cite{cast,deng2021stytr2, Huang2022MGAD} and videos~\cite{mccnet_video, mccnet_v2_video} to achieve style transfer have made great progress. Currently, video style transfer is limited to specific style maps. In this work, we attempt to make real-world video-style manipulation more flexible and creative.

Breakthroughs in the field of text-driven image generation have been made with the public availability of hundreds of millions of large-scale multimodal datasets~\cite{clip, laion5b}. Text-conditional generation efforts, such as DALL-E 2~\cite{dalle2}, Imagen~\cite{imagen}, and Stable Diffusion~\cite{latentdiffusion}, have both powerful image generation capabilities and enhanced user-friendliness, allowing even novices to generate high-quality images. 
It opens up a multitude of possibilities for editing real-world visual videos from current generative works. 
Yet, introducing diffusion models for editing real-world videos remains hard and demanding.
% However, direct use of diffusion models for editing real videos is still challenging.

% 2
Because of the challenges in collecting large-scale data matching to text and video, advancement in the video domain is limited in comparison to that in the image domain.
Training text-guided video generation paradigms~\cite{imagenvideo,makeavideo} from scratch is a time-consuming and resource-intensive task, and tough to acquire and generalize.
% There is yet no generic text-video training paradigm~\cite{imagenvideo,makeavideo} that is publicly available.
% Thus, it is more practical to generate videos using existing text-image models.
Therefore, leveraging current text-image models to generate videos is more practical.
% Consequently, the majority of the current work takes the use of prior generative image models. 
Meanwhile, studies~\cite {vdm,makeavideo,runway,imagenvideo,magicvideo,text2live,tuneavideo} have been implemented on the basis of text-image models for video aspects with promising achievements. 
Text2LIVE~\cite{text2live} propagates editors through the computing explicit correspondences according to pre-trained Neural Layered Atlases(NLA)~\cite{nla} models. 
Tune-A-Video~\cite{tuneavideo} fine-tunes each input video on top of the existing image generation model. 
Despite these efforts, which have contributed to a reduction in resource waste, there is still a substantial computing and timing cost associated with training NLA and fine-tuning input. 
In contrast, we would like to achieve quick inference for arbitrary videos to circumvent the expensive computational consumption.

% The development in the video domain is limited compared to the image domain because of the lack of easy collection of large-scale data corresponding to text and video. Training text-guided video generation models from scratch are resource-intensive, so most of the current work makes use of existing generative image models. Some recent works~\cite{vdm,makeavideo,runway,imagenvideo,magicvideo,text2live,tuneavideo} have further fine-tuned the text-guided image generation model with good results. Text2Live~\cite{text2live} propagates editors by computing explicit correspondences. Tune a video~\cite{tuneavideo} to fine-tune each input video on top of the existing image generation model. These efforts reduce the waste of resources to some extent, but the correspondence computation and fine-tuning costs are still high. We would like to achieve fast inference for arbitrary videos to circumvent the expensive computational costs.

% 3
Furthermore, due to the destructive nature of the noise addition process to the content in the diffusion model and the random character of the denoising process, the input video content is frequently tough to retain.
Moreover, the text prompt in the text-video model only tells the model what the user intends to change, but does not convey what the user prefers to retain. Thus, the text-guided video transfer effort leads to content modifications in the video~\cite{tuneavideo,runway}, which hardly meets the basic requirements of the stylization task~\cite{gatyscnnstyle,avs_2, mccnet_video}.

% Moreover, the input video content is often difficult to retain due to the destructive nature of the noise addition process to the content in the diffusion model, versus the random nature of the denoising process. In addition, the edit text cue in the text-video model only tells the model what the user wants to change, but does not convey what the user intends to keep. Thus, the text-guided video transfer effort leads to content changes in the video~\cite{tuneavideo,runway}, which hardly meets the basic requirements of the stylization task.

% 4
To accomplish stylistic representation of text prompt, preservation of input video content, inter-frame consistency, and fast optimization of the inference process, we concentrate on the following essential components.
First, we propose a new combination of control conditions including text, video frames, and attention maps; specifically, text for style guidance, video frames for content guidance, and attention maps for detail guidance.
And we adapt the noise prediction process by a custom guidance method, which is combined with classifier-free guidance. The global content of the input video is well maintained while realizing the text-guided stylization.
We achieve fast convergence during inference without fine-tuning or additional training, enabling arbitrary stylization tasks for text-video input.
Finally, temporal consistency is also introduced to eliminate flicker and enhance temporal coherence.

% We focus on the following key components to achieving the preservation of the input video content and the optimization of the inference process.
% First, we propose a new combination of control conditions, including text, video frame, and attention map; specifically, the text is used for style guidance, video frame for content guidance, and attention map for detailed structure guidance. The global content of the input video is well maintained while realizing the text-guided stylization.
% Second, we tune the noise prediction process by a custom guidance method, which is inspired by the unclassified guidance, to achieve optimization in the inference process to achieve an arbitrary text-video generation process without fine-tuning or additional training. The optimization method of temporal coherence is also proposed to enhance temporal coherence.

% 5
In summary, we present the following contributions:
\begin{itemize}
\item In this work, we propose \shortname, a novel framework for arbitrary text-driven video styling based on a diffusion model. This work is performed entirely in inference time without additional per-video training or fine-tuning.
\item Novel noise prediction guiding formulas are proposed to achieve simultaneous control of style, content, and structure. Besides, we achieve the control of time and content consistency in the inference process.
\item Various experiments and user studies demonstrate the higher visual quality and effectiveness of our method among corresponding baselines.
% \item User studies and experiments demonstrate that our method is more appealing and effective than corresponding baselines.
\end{itemize}

%% file: Sections/2_related.tex
\section{RELATED WORK}
\label{sec:related_work}
%-------------------------------------------------------------------------
\textbf{Image and video style transfer.}
Image style transfer has been widely studied in recent years, which enables the generation of artistic paintings without the expertise of a professional painter. Gatys \etal~\cite{gatyscnnstyle} find that the inner products of the feature maps in CNNs can be used to represent styles and propose a neural style transfer (NST) method through successive optimization iterations. However, the optimization process is time-consuming and difficult to be widely used. 

A number of techniques~\cite{huangadain,li2017universal,deng2021stytr2,cast} align the second-order statistics of the style and content images to transfer styles arbitrarily. 
Adopting adaptive instance normalization (AdaIN), which normalizes content features using the mean and variance of style features, Huang \etal~\cite{huangadain} present this arbitrary style transfer approach.
To produce domain-specific sequences for content and style, respectively, Deng \etal~\cite{deng2021stytr2} introduce StyTr$2$, which has two separate transformer encoders. 
By comparing and contrasting various styles and taking into account the style distribution, Zhang \etal~\cite{cast} present contrastive arbitrary style transfer (CAST), which enables users to learn style representations directly from image attributes.

% A number of methods~\cite{huangadain,li2017universal,deng2021stytr2,cast} achieve arbitrary style transfer by aligning the second-order statistics of style and content images. 
% Huang \etal~\cite{huangadain} proposed an arbitrary style transfer method by adopting adaptive instance normalization (AdaIN), which normalizes content features using the mean and variance of style features.
% Deng \etal~\cite{deng2021stytr2} propose StyTr$^2$ that contains two different transformer encoders to generate domain-specific sequences for content and style, respectively. 
% Zhang \etal~\cite{cast} present contrastive arbitrary style transfer (CAST) to learn style representation directly from image features by analyzing the similarities and differences between multiple styles and taking the style distribution into account.

The majority of image-style transfer methods~\cite{avs_1,avs_2,avs_3} are employed for video-style transfer. 
The accuracy of the optical flow computation had a significant impact on the sequence-based approaches' effectiveness. Ghosting artifacts will result from using incorrect optical flows. Moreover, processing high-resolution or lengthy videos is challenging due to the computationally expensive nature of predicting optical fluxes. 
We intend to suggest a productive zero-shot video style transfer method that succeeds in both temporal consistency and stylistic expression well.
% Most video style transfer methods rely on the existing image style transfer methods~\cite{avs_1,avs_2,avs_3}. 
% The effect of the sequence-based methods highly depended on the accuracy of optical flow calculation. Using inappropriate optical flows will cause ghosting artifacts. Moreover, estimating optical flows is computationally expensive, which makes processing high-resolution or long videos difficult. 
% We plan to propose an efficient zero-shot video style transfer method that does not require a balance between temporal consistency and stylization effects.

Video style transfer takes a reference style image and statistically applies its style to an input video~\cite{avs_1,avs_2,avs_3,mccnet_video,mccnet_v2_video}. 
In comparison, our method applies a mix of style and content from an input text prompt or image while being constrained by the extracted structure data. By learning a generative model from data, our approach produces semantically consistent outputs instead of matching feature statistics.

\input{Figures/pipeline}
%-------------------------------------------------------------------------
\textbf{Text-to-Image synthesis and editing.}
There has been remarkable progress since the use of conditional GANs in both text-guided image generation~\cite{manigan,dalle}.
ManiGAN~\cite{manigan} contains a text-conditioned GAN for editing an object’s appearance while preserving the image content. However, such multi-modal GAN-based methods are restricted to specific image domains and limited in the expressiveness of the text. DALL-E~\cite{dalle} addresses this by learning a joint image-text distribution over a massive dataset. 
DALL-E produces text-to-images with remarkable accuracy, although it is not intended to edit already-existing images. 

On the other hand, Denoising Diffusion Probabilistic Models (DDPMs)~\cite{ddpm} are successfully leveraged for text-to-image generation. 
GLIDE~\cite{glide} takes this approach further, supporting both text-to-image generation and inpainting.
DALL-E 2~\cite{dalle2} leverages the CLIP~\cite{clip} latent space and a prior model. 
To increase efficiency, Stable Diffusion~\cite{latentdiffusion} generates text from images in latent space rather than pixel space.

A recent influx of approaches makes use of a pre-trained generator~\cite{StyleCLIP,StyleGANNADA} and a pre-trained CLIP~\cite{clip} to provide textual guidance during the generation process.  
StyleCLIP~\cite{StyleCLIP}, StyleGAN-NADA~\cite{StyleGANNADA}, CLIPstyler~\cite{CLIPstyler}, and LDAST~\cite{fu2022ldast} modify the content images based on CLIP optimizations.  
Since StyleCLIP and StyleGAN-NADA are limited by pre-trained generators and can only operate on specific data domains. 
CLIPstyler and LDAST can perform arbitrary content images, however, they still fall short in stylized representation. With the rapid development of generative work, diffusion methods have taken stylization a step further.
Multiple works based on diffusions such as MGAD~\cite{Huang2022MGAD}, DiffStyler~\cite{diffstyler}, and InST~\cite{zhang2022inversion} present impressive results and broaden application scenarios.

%-------------------------------------------------------------------------
\textbf{Text-to-video synthesis and editing.}
Although text-to-image generation has made notable strides, text-to-video generation is still an emerging area of study, primarily because there aren't ample pairs of high-quality text and video.
By mapping text tokens to video tokens, GODIVA~\cite{wu2021godiva} is the first project to extend VQ-VAE~\cite{vqvae} to text-to-video creation. 
N{\"u}wa~\cite{wu2022nuwa} suggests an integrated auto-regressive framework to handle tasks for generating text into both images and videos. 
CogVideo~\cite{hong2022cogvideo} extends CogView-2~\cite{ding2022cogview2} to text-to-video generation utilizing pre-trained text-to-image models and temporal attention modules to increase the quality of the video generation. 
A factorized space-time U-Net is suggested by Video Diffusion Models (VDM)~\cite{vdm} to carry out the diffusion process directly on pixels. 
For the purpose of producing high-resolution videos, Imagen Video~\cite{imagenvideo} has more recently enhanced VDM using cascaded diffusion models and v-prediction parameterization.
Similar in intent, Make-A-Video~\cite{makeavideo} intends to use considerable advancements in text-to-image generation to text-to-video generation. 
They blend the world movements from unsupervised video footage with the appearance-text information from text-image data.

% Although remarkable progress has been achieved in the text-to-image generation, text-to-video generation is a relatively new research field mainly due to the sparsity of high-quality text-video pairs. 
% GODIVA~\cite{wu2021godiva} is the first work to extend VQ-VAE~\cite{vqvae} to text-to-video generation by mapping text tokens to video tokens. 
% N{\"u}wa~\cite{wu2022nuwa} proposes a unified auto-regressive framework to support both text-to-image and text-to-video generation tasks. 
% To improve the video generation quality, CogVideo~\cite{hong2022cogvideo} extends CogView-2~\cite{ding2022cogview2} to text-to-video generation using temporal attention modules and pre-trained text-to-images models. 
% Video Diffusion Models (VDM)~\cite{vdm} proposes a factorized space-time U-Net to perform the diffusion process directly on pixels. 
% More recently, Imagen Video~\cite{imagenvideo} improves VDM with cascaded diffusion models and v-prediction parameterization to generate high-definition videos.
% Make-A-Video~\cite{makeavideo} shares similar motivation and aims to transfer the significant progress from text-to-image generation to text-to-video generation. 
% They combine the appearance-text information from text-image data together with the world movements from unsupervised video footage.

The rapid advancement of text-guided video editing~\cite{text2live,nikankin2022sinfusion,tuneavideo,runway} followed afterward.
By breaking down a video into neural layers, Text2LIVE~\cite{text2live} enables the alteration of input movies using text instructions. 
A layered video representation offers constant propagation across frames once it is accessible. 
By fine-tuning a diffusion model on a single video, SinFusion~\cite{nikankin2022sinfusion} is able to produce variations and extrapolations of videos. 
Similar to this, Tune-A-Video~\cite{tuneavideo} optimizes a video generated from an image model on a single video to allow for modification. However, the viability of these approaches in creative tools is constrained by pricey per-video training. 
Recently, Esser et al. introduce Gen-1~\cite{runway}, which can edit any video with the use of text, however, the edited version loses the input video's content information to some extent.

% Text2Live~\cite{text2live} allows editing input videos using text prompts by decomposing a video into neural layers. 
% Once available, a layered video representation provides consistent propagation across frames. 
% SinFusion~\cite{nikankin2022sinfusion} can generate variations and extrapolations of videos by optimizing a diffusion model on a single video. 
% Similarly, Tune-A-Video~\cite{tuneavideo} finetunes an image model converted to video generation on a single video to enable editing. However, expensive per-video training limits the practicality of these approaches in creative tools. 
% Recently, \cite{runway} proposed by Esser \etal is able to modify arbitrary videos under the guidance of the text, however, the editing result loses the content information of the input video.

%% file: Figures/pipeline.tex
\begin{figure*}%[tbp]
    \centering
    \includegraphics[width=0.85\linewidth]{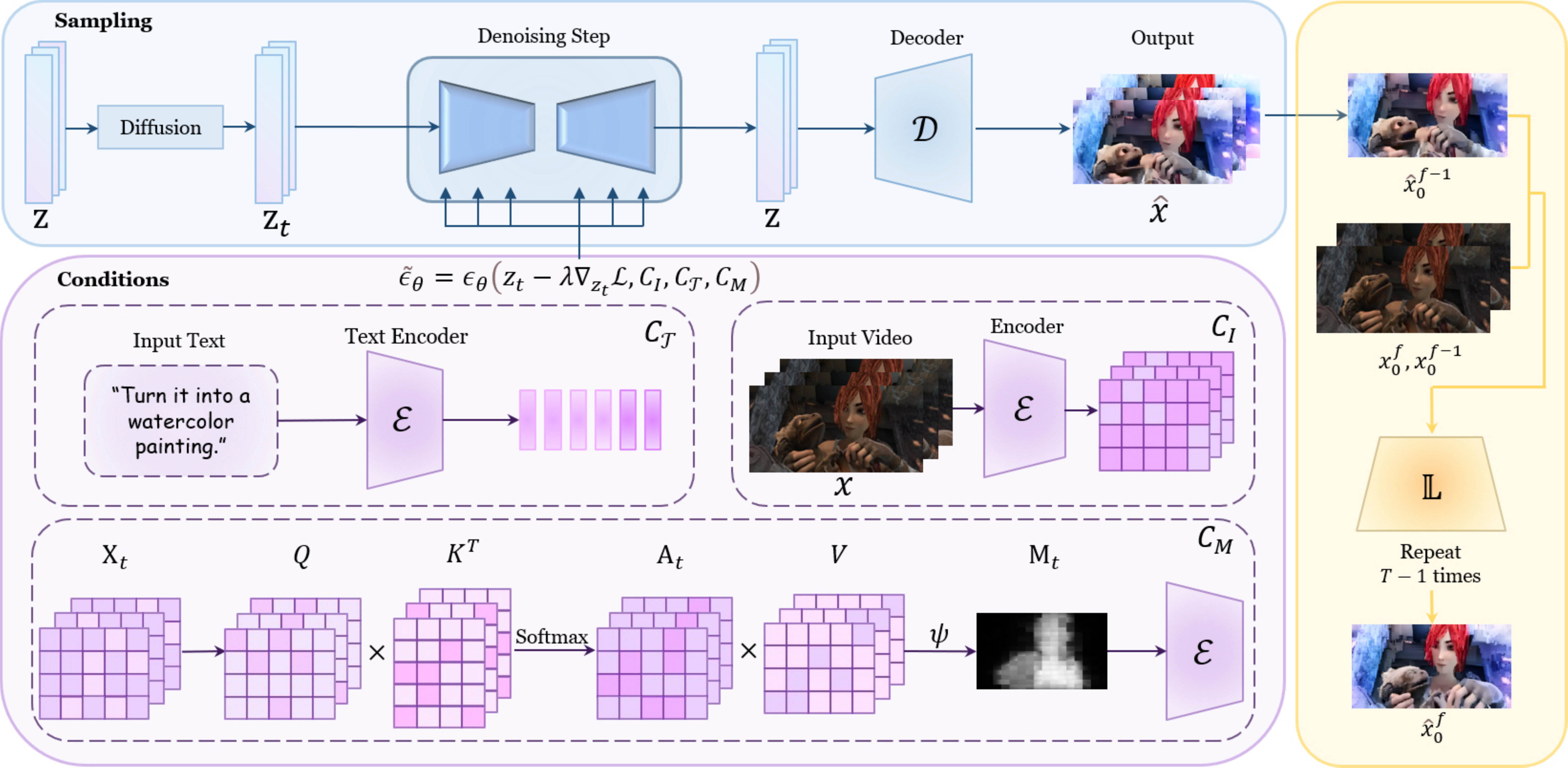}
    % \vspace{-0.2cm} 
    \caption{The overall framework of our method for text-driven video style transfer. The purple, blue, and yellow parts are the conditional representation, the sampling module, and the temporal consistency module, respectively. It generates the stylized video by maintaining the input video content while transferring the style through the textual description of the content.}
    % \vspace{-0.4cm} 
    \label{fig:pipeline}
\end{figure*}

%% file: Sections/3_method.tex
\section{METHODS}
\label{sec:method}
The purpose of our proposed approach is to stylize the input video for editing according to the text condition $\mathcal{T}$ while preserving the content and structure of the input video.
Content is specifically defined as the appearance and semantic information of the input video. Structure refers to the shape and geometric features of the input video.
To achieve that, we add the input frame information (denoted by $I$) with the self-attention information (denoted by $M$) to the generative model conditions. We infer the self-attention map $M$ between encoded text $\mathcal{T}$ and intermediate features of the denoiser $\epsilon_\theta$.
The specific procedure is shown in Fig.~\ref{fig:pipeline}. 
First, the implementation of our generative backbone is formulated as a conditional potential video diffusion model. 
Next, the guidance conditions are computed and encoded. 
Then, the guidance conditions and the classifier-free guidance are combined for noise prediction. 
% First, we describe our implementation of the generative model as a conditional potential video diffusion model. 
% Next, we compute and encode the guidance conditions. 
% Then, we combine the guidance conditions with the classifier-free guidance for noise prediction. 
And step-wise optimization is performed during the sampling process. 
Finally, the temporal consistency between frames is optimized.

% The specific process is as follows. First, we describe our implementation of the generative model as a conditional potential video diffusion model, and then, we compute self-attention between encoded text and intermediate features of the denoiser. Then, we combine the new guidance condition with the classifier-free guidance for noise prediction. Finally, we discuss the optimization process of our model.

%---------------------------------------------------------DM---------------------------------------------------------
\subsection{Diffusion Models}
% \textbf{【runway】}，【pix2pix zero】
Diffusion models~\cite{2015Diffusion} learn to reverse a fixed forward diffusion process.
% , which is defined as 
% \begin{equation}
% q\left(x_t \mid x_{t-1}\right):=\mathcal{N}\left(x_t, \sqrt{1-\beta_t} x_{t-1}, \beta_t \mathbb{I}\right).
% \end{equation}
Normally-distributed noise is slowly added to each sample $x_{t-1}$ to obtain $x_{t}$. The forward process models a fixed Markov chain and the noise is dependent on a variance schedule $\beta_t$ where $t \in\{1, \ldots, T\}$, with T being the total number of steps in our diffusion chain, and $x_0 := x$.

%---------------------------------------------------------DDPM---------------------------------------------------------
\textbf{Denoising Diffusion Probabilistic Models (DDPMs)}~\cite{ddpm} are latent generative models trained to recreate a fixed forward Markov chain $x_1, \ldots, x_T$. Given the data distribution $x_0 \sim q\left(x_0\right)$, the Markov transition $q\left(x_t \mid x_{t-1}\right)$ is defined as a Gaussian distribution with a variance schedule $\beta_t \in(0,1)$, that is,
\begin{equation}
q\left(x_t \mid x_{t-1}\right)=\mathcal{N}\left(x_t ; \sqrt{1-\beta_t} x_{t-1}, \beta_t \mathbb{I}\right),
\end{equation}
where $t=1, \ldots, T$.
By the Bayes’ rules and Markov property, one can explicitly express the conditional probabilities $q\left(x_t \mid x_{0}\right)$ and $q\left(x_{t-1} \mid x_t, x_0\right)$ as
\begin{equation}
% \begin{array}{r}
q\left(x_t \mid x_0\right)=\mathcal{N}\left(x_t ; \sqrt{\bar{\alpha}_t} x_0,\left(1-\bar{\alpha}_t\right) \mathbb{I}\right), 
\end{equation}
\begin{equation}
q\left(x_{t-1} \mid x_t, x_0\right)=\mathcal{N}\left(x_{t-1} ; \tilde{\mu}_t\left(x_t, x_0\right), \tilde{\beta}_t \mathbb{I}\right),  
\end{equation}
\begin{equation}
\begin{array}{r}
\text { w.r.t. }  t=1, \ldots, T,  \alpha_t=1-\beta_t, \\
\bar{\alpha}_t=\prod_{s=1}^t \alpha_s, \tilde{\beta}_t=\frac{1-\bar{\alpha}_{t-1}}{1-\bar{\alpha}_t} \beta_t, \\
\tilde{\mu}_t\left(x_t, x_0\right)=\frac{\sqrt{\bar{\alpha}_t} \beta_t}{1-\bar{\alpha}_t} x_0+\frac{\sqrt{\alpha_t}\left(1-\bar{\alpha}_{t-1}\right)}{1-\bar{\alpha}_t} x_t .
\end{array}
\end{equation}

To generate the Markov chain $x_1, \ldots, x_T$, DDPMs leverage the reverse process with a prior distribution $p(x_T) = \mathcal{N}\left(x_T ; 0, \mathbb{I}\right)$ and Gaussian transitions
\begin{equation}
p_\theta\left(x_{t-1} \mid x_t\right)=\mathcal{N}\left(x_{t-1} ; \mu_\theta\left(x_t, t\right), \Sigma_\theta\left(x_t, t\right)\right),
\end{equation}
where $t \in\{T, \ldots, 1\}$.
Learnable parameters $\theta$ are trained to guarantee that the generated reverse process is close to the forward process.

To this end, DDPMs follow the variational inference principle by maximizing the variational lower bound of the negative log-likelihood, which has a closed form given the KL divergence among Gaussian distributions. Empirically, these models can be interpreted as a sequence of weight-sharing denoising autoencoders $\epsilon_\theta\left(x_t, t\right)$, which are trained to predict a denoised variant of their input $x_t$. The objective can be simplified as 
\begin{equation}
\mathbb{E}_{x, \epsilon \sim \mathcal{N}(0,1), t}\left[\left\|\epsilon-\epsilon_\theta\left(x_t, t\right)\right\|_2^2\right].
\end{equation}

\input{Figures/moreresult}
\input{Alg/alg1}

%---------------------------------------------------------LDM---------------------------------------------------------
\textbf{Latent Diffusion Models (LDMs)}~\cite{latentdiffusion}
% \textbf{【instruct pix2pix 3.2】}
are newly introduced variants of DDPMs that operate in the latent space of an autoencoder. 
LDMs improve the efficiency and quality of diffusion models by operating in the latent space of a pre-trained variational autoencoder~\cite{vae}. 
LDMs consist of two key components. First, the autoencoder is trained with patch-wise losses on a large collection of images.
For an image $x$, the diffusion process adds noise to the encoded latent $z = \mathcal{E}(x)$ producing a noisy latent $z_t$ where the noise level increases over timesteps $t \in T$. 
And a decoder $\mathcal{D}$ learns to reconstruct the latent back to pixel space, such that $\mathcal{D}(\mathcal{E}(x)) \approx x$. The second component is a DDPM that is trained to remove the noise added to a latent representation of an image. This diffusion model can be conditioned on encoded embeddings of class labels.
We learn a network $\theta$ that predicts the noise added to the noisy latent $z_t$ given image condition $c_I$, text prompt condition $c_\textprompt$, and attention map condition $c_M$.
We minimize the following latent diffusion objective:
\begin{equation}
\left.L=\mathbb{E}_{\mathcal{E}(x), c_I, c_\textprompt, c_M, \epsilon \sim \mathcal{N}(0,1), t}\left[\| \epsilon-\epsilon_\theta\left(z_t, t, c_I, c_\textprompt, c_M\right)\right) \|_2^2\right].
\end{equation}

\subsection{Condition Representation}
\textbf{Style condition.}
Conditional Diffusion Models are well-suited to modeling conditional distributions such as $p(x \mid c)$. In this case, the forward process $q$ remains unchanged while the conditioning variables $c$ become additional inputs to the model. Our goal is to edit an input video based on a text prompt describing the desired edited video.
Therefore, during sampling, we replace the category condition with the textual prompt description. The target style  can be obtained from the textual prompts $c_\textprompt$ and can be represented by the following equation:
\begin{equation}
z \sim p_\theta(z \mid c_\textprompt), \quad x=\mathcal{D}(z).
\end{equation}

\textbf{Content condition.}
For the stylization task, besides style representation, content retention is another key issue. Due to the destructive nature of the noise addition process of the diffusion model on the content map itself, as well as the random nature of the denoising process, it is difficult to achieve better results on content retention. Previous works~\cite{runway,balaji2022ediffi}, which used CLIP~\cite{clip} image embedding to represent the content conditions, resulted in difficulties in achieving the traditional stylization requirements.
In contrast, we add additional input channels to the first convolutional layer to concatenate $z_t$ and $c_I$, so that the final generated results have more consistency in semantic content relative to the input video.
The pre-trained checkpoints are used to initialize all of the diffusion model's available weights, and the weights that act on the newly added input channels are set to zero.

% Content Representation To infer a content representation from both text inputs t and video inputs x, we follow previous works [35, 3] and utilize CLIP [32] image embeddings to represent content. For video inputs, we select one of the input frames randomly during training. Similar to [35, 49], one can then train a prior model that allows sampling image embeddings from text embeddings. This approach enables specifying edits through image inputs instead of just text. Decoder visualizations demonstrate that CLIP embeddings have increased sensitivity to semantic and stylistic properties while being more invariant towards precise geometric attributes, such as sizes and locations of objects [34]. Thus, CLIP embeddings are a fitting representation of content as structure properties remain largely orthogonal.

\textbf{Self-features condition.}
% \subsection{Cross-attention in conditioned Diffusion Models}
% 【prompt to prompt】【SAG】
%[zero pix2pix]  (改的时候可以参考一下prompt to prompt）
% Recent large-scale diffusion models~\cite{Huang2022MGAD, glide,prompt2prompt} incorporate conditioning by augmenting the denoising network $\epsilon_\theta$ with the cross-attention layer~\cite{prompt2prompt, sag}. We use the open-source Stable Diffusion model, built on LDMs~\cite{latentdiffusion}. The model produces text embedding $c_\textprompt$ with the CLIP~\cite{clip} text encoder. Next, to condition the generation on text, the model computes cross-attention between encoded text and intermediate features of the denoiser $\epsilon_\theta$:
% \begin{equation}
% \operatorname{Attention}(Q, K, V)=M \cdot V.
% \end{equation}
Recent large-scale diffusion models~\cite{Huang2022MGAD, glide,prompt2prompt} incorporate conditioning by augmenting the denoising U-Net $\epsilon_\theta$ with the attention layer~\cite{prompt2prompt, sag}.
The self-attention mechanism is a variant of the attention mechanism, which relies less on external information and is better at capturing the internal relevance of self-features.
Specifically, given any feature map $X_t \in \mathbb{R}^{(HW) \times d}$ at a timestep $t$, for the height $H$ and width $W$, the $N$-head self-attention is defined as:
\begin{equation}
Q_t^{(h)}=X_t W_Q^{(h)}, \quad K_t^{(h)}=X_t W_K^{(h)},
\end{equation}
where $W_Q^{(h)}, W_K^{(h)} \in \mathbb{R}^{C \times d}$ for $h=0,1, \ldots, N-1$. 
\begin{equation}
A_t^{(h)}=\operatorname{softmax}\left(Q_t^{(h)}\left(K_t^{(h)}\right)^T / \sqrt{d}\right).
\end{equation}
Each $A_t^{(h)}$ is then right multiplied by $V_t^{(h)}=X_t W_V^{(h)}$ where $W_V^{(h)} \in \mathbb{R}^{C \times d}$.
Given a masking threshold $\psi$ which is practically set to the mean value of $A_t$, the masked patches of $x_t$ according to the self-attention map, and is formulated as follows:
\begin{equation}
M_t=\mathbbm{1} \left(A_t>\psi\right)
\end{equation}

% Recent large-scale diffusion models~\cite{Huang2022MGAD, glide,prompt2prompt} incorporate conditioning by augmenting the denoising network $\epsilon_\theta$ with the cross-attention layer~\cite{prompt2prompt, sag}. We use the open-source Stable Diffusion model, built on LDMs~\cite{latentdiffusion}. The model produces text embedding $c_\textprompt$ with the CLIP~\cite{clip} text encoder. Next, to condition the generation on text, the model computes cross-attention between encoded text and intermediate features of the denoiser $\epsilon_\theta$:
% \begin{equation}
% \operatorname{Attention}(Q, K, V)=M \cdot V.
% \end{equation}
% Query $Q=W_Q \varphi\left(x_t\right)$, key $K=W_K c$, and value $V=W_V c$ are computed with the learnt projections $W_Q$, $W_K$, $W_V$ applied on intermediate spatial features $\varphi\left(x_t\right)$ of the denoising UNet~\cite{unet} $\epsilon_\theta$ and the text embedding $c_\textprompt$, and $d$ is the dimension of projected keys and queries.
% The attention maps are then 
% \begin{equation}
%  M=\operatorname{Softmax}\left(\frac{Q K^T}{\sqrt{d}}\right).
% \end{equation}
% Of particular interest is the cross-attention map $M$, which is observed to have a tight relation with the structure of the image~\cite{prompt2prompt}.

\input{Figures/comparison_tune}

\subsection{Condition Guidance}
% 【instruct pix2pix 3.2.1】【runway-Content Representation，Conditioning Mechanisms】【SAG】
\textbf{Classifier-free guidance}~\cite{ho2022classifier} 
% 【instruct pix2pix 3.2.1】
is a method for trading off the quality and diversity of samples generated by a diffusion model. It is commonly used in class-conditional and text-conditional image generation to improve the visual quality of generated images and to make sampled images better correspond with their conditioning. Classifier-free guidance effectively shifts probability mass toward data where an implicit classifier $p_\theta\left(z_t \mid c\right)$ assigns a high likelihood to the conditioning $c$. The implementation of classifier-free guidance involves jointly training the diffusion model for conditional and unconditional denoising, and combining the two score estimates at inference time. Training for unconditional denoising is done by simply setting the conditioning to a fixed null value $c=\varnothing$ at some frequency during training. At inference time, with a guidance scale $s \geq 1$, the modified score estimate $\tilde{\epsilon_\theta}\left(z_t, c\right)$ is extrapolated in the direction toward the conditional $\epsilon_\theta\left(z_t, c\right)$ and away from the unconditional $\epsilon_\theta\left(z_t, \varnothing\right)$:
\begin{equation}
\tilde{\epsilon_\theta}\left(z_t, c\right)=\epsilon_\theta\left(z_t, \varnothing\right)+s \cdot\left(\epsilon_\theta\left(z_t, c\right)-\epsilon_\theta\left(z_t, \varnothing\right)\right).
\end{equation}
% 如果字数不够的话，看classifier-free的Appendix B

\textbf{Guidance for conditions.}
% 【instruct pix2pix】
For our task, the scoring network $\tilde{\epsilon_\theta}\left(z_t, c_I, c_\textprompt\right)$ has three conditions: the input image $c_I$, text prompt $c_\textprompt$, and self-attention map $c_M$. We find it beneficial to leverage classifier-free guidance:
\begin{equation}
\epsilon^{\prime}_\theta=\epsilon_\theta\left(z_t, \varnothing, \varnothing\right)
\end{equation}
concerning each condition.
Liu \etal~\cite{liu2022compositional} demonstrate that a conditional diffusion model can compose score estimates from multiple different conditioning values. We introduce three guidance scales, $s_I$, $s_\textprompt$, and $s_M$, which can be adjusted to trade off how strongly the generated samples correspond with the conditions. Our modified score estimate is as follows:
\begin{equation}
\begin{aligned}
\tilde{\epsilon_\theta}\left(z_t, c_I, c_\textprompt, c_M\right) = &(1-s_I-s_\textprompt-s_M)\cdot \epsilon^{\prime}_\theta \\
& + s_I \cdot\epsilon_\theta\left(z_t, c_I, \varnothing\right) \\
&+ s_\textprompt \cdot\epsilon_\theta\left(z_t, \varnothing, c_\textprompt\right) \\
& + s_M \cdot\epsilon_\theta\left(z_t, c_M, \varnothing\right).
\end{aligned}
\end{equation}

% \begin{equation}
% \begin{aligned}
% \tilde{\epsilon_\theta}\left(z_t, c_I, c_\textprompt, c_M\right)= & \epsilon_\theta\left(z_t, \varnothing, \varnothing\right) \\
% & +s_I \cdot\left(\epsilon_\theta\left(z_t, c_I, \varnothing\right)-\epsilon_\theta\left(z_t, \varnothing, \varnothing\right)\right) \\
% & +s_\textprompt \cdot\left(\epsilon_\theta\left(z_t, \varnothing, c_\textprompt\right)-\epsilon_\theta\left(z_t, \varnothing, \varnothing\right)\right) \\
% & +s_M \cdot\left(\epsilon_\theta\left(z_t, c_M, \varnothing\right)-\epsilon_\theta\left(z_t, \varnothing, \varnothing\right)\right).
% \end{aligned}
% \end{equation}

\subsection{Sampling Optimization}
\textbf{Loss.}
% Text2Live
We want to allow substantial texture and appearance changes while preserving the objects’ original spatial layout, shape, and perceived semantics. While various perceptual content losses have been proposed in the context of style transfer, most of them use features extracted from a pre-trained VGG model. Instead, we define our loss in CLIP feature space. This allows us to impose additional constraints on the resulting internal CLIP representation of Io. Inspired by classical and recent works~\cite{tumanyan2022splicing}, we adopt the self-similarity measure. Specifically, we feed an image into CLIP’s ViT encoder and extract its spatial tokens from the deepest layer. The structure loss is denoted by $\mathcal{L}_s$. 
The network is optimized by minimizing the similarity between the input frame ${x}^f_0$ and the predicted frame ${x}^f_t$:
\begin{equation}
\mathcal{L}_s = 1 - \mathcal{D}_{\cos }\left({x}^f_0, {x}^f_t\right),
\end{equation}
where $f \in\{1, \ldots, F\}$, $F$ being the frame number of the input video.
We then take a gradient step according to the $z$ gradient $\nabla_{z_t}$:
\begin{equation}
% \Delta z_t=\nabla_{z_t}\loss_{t o t a l},
\Delta z_t=\nabla_{z_t}\loss_{s},
\end{equation}
which optimizes the denoising network together with the guidance condition, as in the following equation:
\begin{equation}
\tilde{\mathbf{\epsilon}}_\theta = {\mathbf{\epsilon}}_\theta ( z_t - \lambda \Delta z_t, C_I, C_\textprompt, C_M).
\end{equation}
% 【pix2pix zero】，text2live，disentangled

\subsection{Temporal Consistency}
The video frames $\left\{x_0^f\right\}_{f=1}^F$ are consistent with each other globally in both the short term and the long term. However, it might contain local flicker due to the misalignment between input and atlas-based frames. Hence, we use an extra local deflicker network to refine the results further. Prior work has shown that local flicker can be well addressed by a flow-based regularization. Hence, we choose a lightweight pipeline [27] with modification. As shown in Fig~\ref{fig:pipeline}, we predict the output frame $\hat{x}_0^{f}$ by providing two consecutive frames $x_0^f$, $x_0^{f-1}$ and previous output ${\hat{x}_0}^{f-1}$ to local refinement network $\mathbb{L}$. Two consecutive frames are firstly followed by a few convolution layers and then fused with the ${\hat{x}_0}^{f-1}$. 
The local flickering network is trained with a temporal consistency loss to remove local flickering artifacts~\cite{lei2023blind}.
% \begin{equation}
% \mathcal{L}_{\text {local }}\left(x_0^{f}, x_0^{f-1}\right)=\left\|M_{f, f-1} \odot\left(x_0^{f}-\hat{x_0}^{f-1}\right)\right\|_1,
% \end{equation}
% where $\hat{x_0}^{f-1}$ is obtained by warping the ${x_0}^{f-1}$ with the optical flow from frame $t$ to frame $t − 1$. $M_{f,f-1}$ is the corresponding occlusion mask. For the frames without local artifacts, the output should $x_0^{f}$ be the same as $$. Hence, we also provide a reconstruction loss by minimizing the distance between Ot and Of t to regularize the quality.

%% file: Figures/moreresult.tex
\begin{figure*}[thbp]
    \centering
    \includegraphics[width=\linewidth]{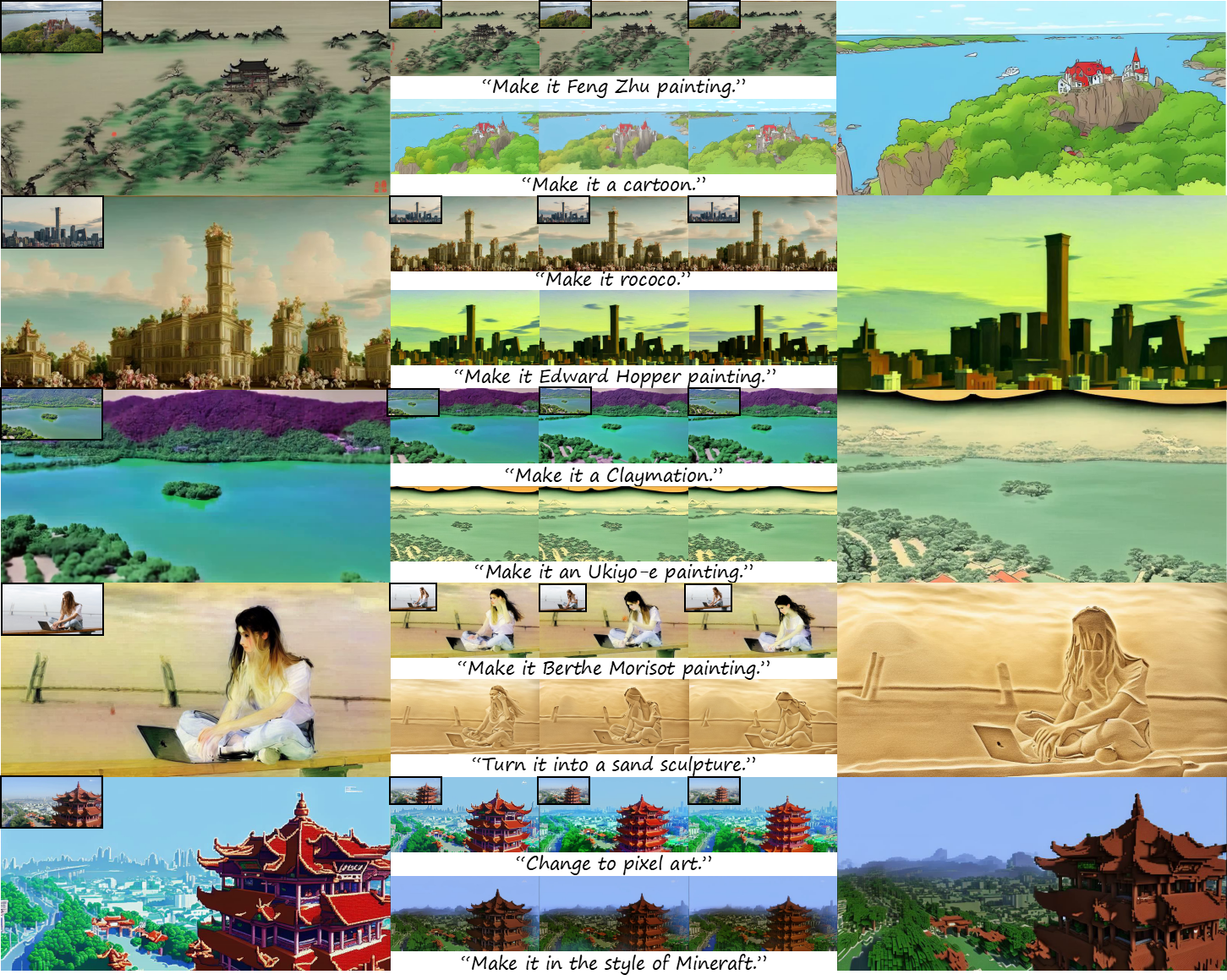}
    % \vspace{-0.2cm} 
    \caption{\textbf{More video style transfer results by Style-A-Video.} Our approach supports simple arbitrary style descriptions, without the necessity to describe the visual content in detail as \cite{tuneavideo,fatezero} require. The input contents for each result are shown as inset.
    }
    % \vspace{-0.4cm} 
    \label{fig:moreresult1}
\end{figure*}

%% file: Alg/alg1.tex
\begin{algorithm}[H]
\caption{Conditions Guidance Diffusion Sampling.}
\begin{algorithmic}
\STATE 
\STATE {\textbf{Input:}} The text prompt $\textprompt$, video frame $I$, \\frame number $F$, Diffusion model Model($z_t$).
\STATE {\textbf{Output:}} Stylized frames ${x^{1}_0,..., x^{F}_0}$.

\STATE  $\mathbf{z}_T \sim \mathcal{N}(0, \mathbf{I})$ a unit Gaussian random variable with specific seed $\mathbf{S}$
% \STATE  $\mathbf{z}_T$ = encode$\mathbf{x}_T$
% \vspace{3mm}

\STATE \hspace{0.25cm} \textbf{for} $f$ in $1,..., F$ \textbf{do}
\STATE \hspace{0.25cm} $x^{f}_t \leftarrow noising(x^{f}_0)$
\STATE \hspace{0.25cm} $z_t = \mathcal{E}(x^{f}_t)$

\STATE \hspace{0.5cm} \textbf{for} $t$ in $T,..., 1$ \textbf{do}
\STATE \hspace{0.85cm} $\epsilon, \Sigma, M \leftarrow \operatorname{Model}\left(\mathbf{z}_t\right)$
% \STATE \hspace{0.85cm} $\loss$ =  ???
\STATE \hspace{0.85cm} $\Delta z_t=\nabla_{z_t}\loss_s$
\STATE \hspace{0.85cm} $C_I, C_\textprompt, C_M \leftarrow$ CLIP embedding $(I, \textprompt, M)$
\STATE \hspace{0.85cm} $\tilde{\mathbf{\epsilon}}_\theta = {\mathbf{\epsilon}}_\theta ( z_t - \lambda \Delta z_t, C_I, C_\textprompt, C_M)$
\STATE \hspace{0.85cm} $\mathbf{z}_{t-1} \sim \mathcal{N}\left(\frac{1}{\sqrt{\bar{\alpha}_t}}\left(\mathbf{z}_t-\frac{1-\alpha_t}{\sqrt{1-\bar{\alpha}_t}} \tilde{\epsilon}\right), \Sigma\right)$
\STATE \hspace{0.5cm} \textbf{end for}
\STATE \hspace{0.5cm} \textbf{return $z_0$}
% \vspace{3mm}
% \STATE  \hspace{0.5cm} $\mathbf{x}_0$ = $\mathcal{D}(\mathbf{z}_0)$

\STATE \hspace{0.5cm} $\hat{x}^{f}_0 = \mathcal{D}(\mathbf{z}_0)$
\STATE \hspace{0.5cm} \textbf{if} $f = 1$
\STATE \hspace{0.85cm} $\hat{x}^{f-1}_0 = \varnothing$
\STATE \hspace{0.5cm} \textbf{else}
\STATE \hspace{0.85cm} $\hat{x}^{f}_0 \leftarrow x^{f}_0, \hat{x}^{f-1}_0$
\STATE \hspace{0.25cm} \textbf{end for}
\STATE \hspace{0.25cm} \textbf{return $x^{f}_0$}
% \vspace{1mm}
\end{algorithmic}
\label{alg1}
\end{algorithm}

%% file: Figures/comparison_tune.tex
\begin{figure*}%[thbp]
    \centering
    % \vspace{-0.8cm} 
    \includegraphics[width=\linewidth]{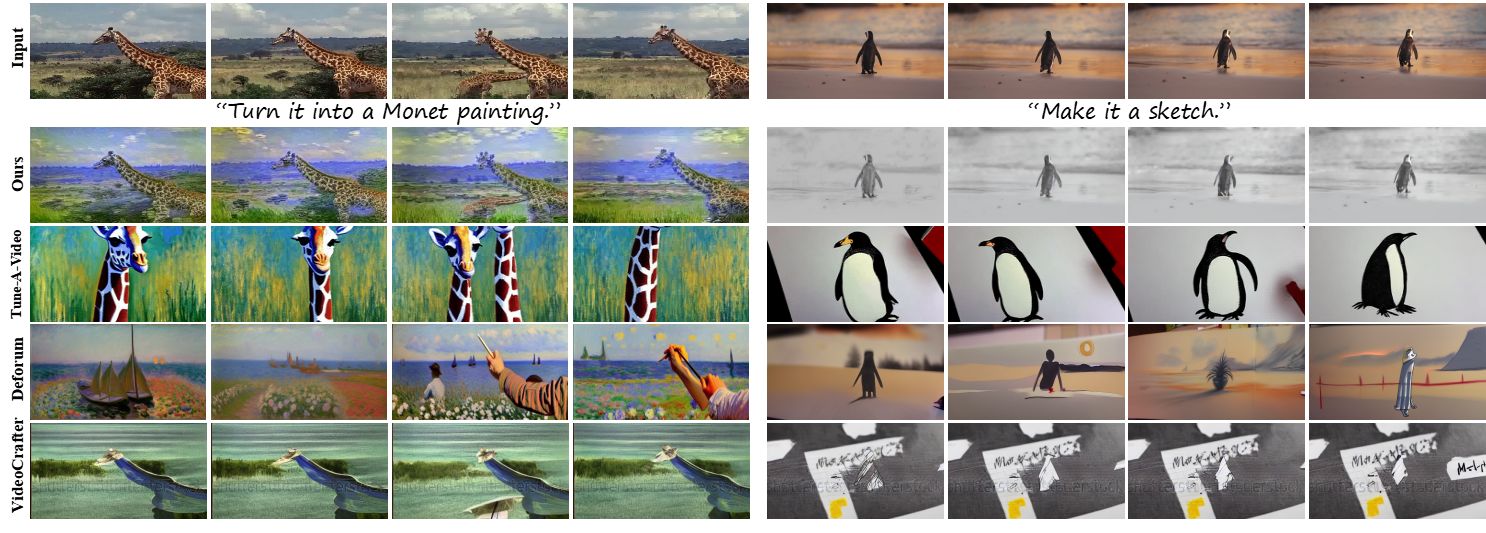}
    % \vspace{-0.4cm} 
    \caption{\textbf{Qualitative comparison of Style-A-Video with other baselines.} Our results have superior temporal consistency, content preservation, and style representation.}
    \label{fig:comparison_tune}
    % \vspace{-0.5cm} 
\end{figure*}

%% file: Sections/4_experiments.tex
\input{Figures/comparison_text2live}

\input{Tables/quantity}
\section{EXPERIMENTS}
\subsection{Implementation Details}
In practice, we implement our approach through latent diffusion models (LDMs)~\cite{latentdiffusion}. Our development is based on the capabilities of two large-scale pre-trained models operating~\cite{brooks2022instructpix2pix} on different modalities - the large-scale language model GPT-3~\cite{gpt3} and the image generation model Stable Diffusion~\cite{latentdiffusion}.
Style-A-Video takes $1$ second and consumes $3925$ MiB to generate a $512\times256$ frame on a single NVIDIA GeForce RTX $3090$. 
The U-Net~\cite{unet} architecture we utilize is based on Wide-ResNet~\cite{wideres}. 
To ensure the quality of the results and to maintain consistency of the parameters, the diffusion step and the time step used for the experiments in our work are set to $30$. 

% \subsection{Computing Consumptions}
% Style-A-Video takes $1$ seconds to generate a $512\times256$ frame on a single NVIDIA GeForce RTX $3090$. 
% % And it takes $14$ seconds with a single channel. 
% % This is a moderate amount of time compared to related work.
% % ------------------改------------------begin
% In particular, it is a superior inference time compared to diffusion-related work~\cite{DiffusionCLIP,latentdiffusion}.
% The inference process for one image consumes about 4 GB (6 GB and 16 GB for DiffusionCLIP~\cite{DiffusionCLIP} and Stable Diffusion~\cite{latentdiffusion}, respectively), which is much smaller than the general diffusion model~\cite{ruiz2022dreambooth, imagen}. 
% % ------------------改------------------end

\subsection{Qualitative Evaluation}
\textbf{Results analysis.}
% text2live,rdm,runway
We test our approach on a variety of videos and stylized texts. These videos are sourced from the web and contain various object categories, such as people, animals, landscapes, etc.
Figs.~\ref{fig:teaser} and \ref{fig:moreresult1}  show the results obtained by using different input videos and text prompts to guide.
Our method can handle arbitrary shots, such as large wide-angle landscape shots with close-ups of people. In addition, we can generate stable results for both still shots and moving shots without additional subject tracking.
% As shown in Fig.~\ref{fig:teaser} in the video content of the person, we can stylize the fine facial features and hair of the person. For action clips in movies, stylized results with temporal consistency can also be generated.
% In summary, the results of our method are successful in stylizing arbitrary images.
% In Sec.~\ref{subsec:quantitative}, we also evaluate our method's ability to control other features such as temporal consistency and preservation of the input structure.

\input{Figures/userstudy}
\textbf{Comparison with baselines.}
We compare Style-A-Video with SOTA text-driven video editing methods, including Text2LIVE~\cite{text2live}, Tune-A-Video~\cite{tuneavideo}, Defroum~\cite{deforum}, and VideoCrafter~\cite{videocrafter}. Tune-A-Video requires additional training for each input video and text. The results in Fig.~\ref{fig:comparison_tune} show that Tune-A-Video has difficulty reproducing the content, shape, motion, and position of the input video; for example, the size and pose of the penguin and giraffe do not match the input video. And some content and artifacts are generated that do not match the input content, and from the results, backgrounds are generated that do not match the input content. Deforum generates high quality in the resulting images, however, there is no coherence between frames. In the giraffe example, the giraffe turns into a boat and the arm of a person who is drawing. VideoCrafter wreaks havoc on content in terms of presentation style. Unlike the above baseline, our method does a good job of maintaining the content and motion of the input video during editing, while performing stylistic expressions and achieving a high-quality video stylization.
In addition, we compare our model with the publicly available ``car'' based pre-trained model of Text2LIVE~\cite{text2live}, which maintains the content of the input video rather well but struggles to meet the editing requirements imposed by the text. As demonstrated in Fig.~\ref{fig:comparison_text2live}, Text2LIVE is more limited in terms of style expressiveness than our results.

\subsection{Quantitative Evaluation}
\label{subsec:quantitative}
We conduct the quantitative evaluation using the trained CLIP~\cite{clip} model as previous methods~\cite{tuneavideo,deforum,text2live,videocrafter}. We randomly select 25 videos generated by Style-A-Video and each SOTA method. We quantify the trade-off between time consistency, cue consistency, and frame accuracy using the following three metrics, respectively.

\textbf{Temporal consistency.}
We measure the temporal consistency of frames by first calculating the CLIP image embedding on all frames of the output video and then calculating the cosine similarity between all consecutive frame pairs measured the temporal consistency of frames.

\textbf{Prompt consistency.}
We first calculate the CLIP image embedding on the output frames of the output video and the CLIP text embedding for the stylized text prompt. Besides, we calculate the average cosine similarity between the stylized frame embedding and the text embedding. This is thus used to measure the consistency of text and stylization.

\textbf{Frame accuracy.}
We calculate the CLIP image embedding for the input video frames, with their stylized frames. The average cosine similarity between the two is calculated. This is used to measure the degree of content preservation for the input video. 

As can be seen in Table~\ref{tab:quantity}, the Style-A-Video method achieves superior results in terms of temporal consistency and prompt consistency compared to the baselines and shows comparable content preservation accuracy to the TextLIVE~\cite{text2live} method.

\subsection{User Study}
\textbf{User Study \uppercase\expandafter{\romannumeral1}.}
We compare our approach with several open-source SOTA text-guided video editing methods, including Text2LIVE~\cite{text2live}, VideoCrafter~\cite{videocrafter}, Deforum~\cite{deforum}, and Tune-A-Video~\cite{tuneavideo}. All baselines are trained and reasoned using publicly available implementations with default configurations. We randomly display $20$ sets of content-style pairs, each pair containing the results of Style-A-Video with one of the other methods. We required $70$ participants to select the more favored result and collected 1400 votes. The last row of Table~\ref{tab:quantity} reports the percentage of votes each method received compared to Style-A-Video.

\textbf{User Study \uppercase\expandafter{\romannumeral2}.}
We design a novel and detailed user study to better evaluate the performance of various aspects of the generated results. For each participant, $40$ text-video pairs are randomly provided. We requested volunteers to score the following criteria: ``how well the results agree with the style of the text description'', ``how well the results agree with the content of the input video'', ``temporal consistency of the results'', and ``overall effects''. Finally, we collect $11200$ scores from $70$ participants. Fig.~\ref{fig:userstudy} shows the average scores of each method. It shows that our methods achieved good results on all indicators.

\input{Figures/ablation_study}
\input{Figures/ablation_wo}
\subsection{Ablation Study}
Fig.~\ref{fig:ablation} shows the effect of each guidance condition $C_I$, $C_\textprompt$, and $C_M$ on the model we proposed.
Larger guidance coefficients $s_I$, $s_\textprompt$, and $s_M$ imply an increase in guidance weights. We observe a slight trade-off for increasing the intensity parameter in the baseline model. A larger $s_I$ implies stronger agreement with the input video but reduces agreement with the style guide. Similarly, a larger $s_\textprompt$ implies a stronger stylization effect and thus more deviation from the content of the input video.

To assess the impact of the self-attention mask condition and temporal consistency module on the Style-A-Video findings, we conduct an ablation study. As shown in Fig.~\ref{fig:ablation_wo}, each design is ablated individually to analyze its impact. The second column investigates the effectiveness of the self-attention mask condition. Removal of the self-attention map condition leads to the loss of fine details (e.g., towns, temples, wings). The third column investigates the effect of the temporal consistency module. In the absence of this module, differences in color and illumination are produced between frames, resulting in the flickering of the video. These results show that all the above key designs contribute to the successful results of our approach.

%% file: Figures/comparison_text2live.tex
\begin{figure}%[thbp]
    \centering
    % \vspace{-0.8cm} 
    \includegraphics[width=\linewidth]{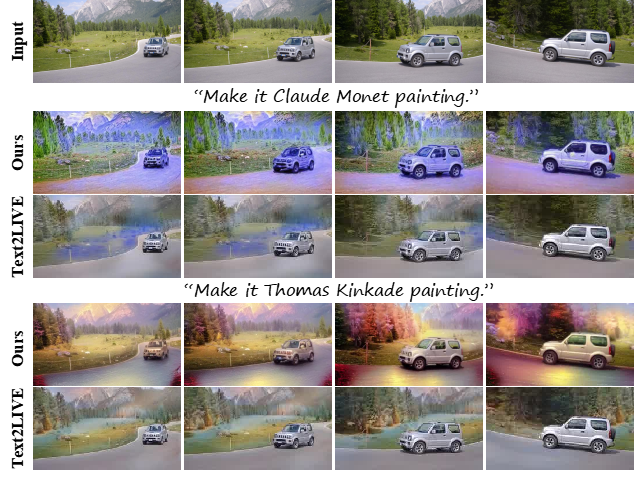}
    % \vspace{-0.4cm} 
    \caption{\textbf{More qualitative comparison.} We conducted it utilizing the pre-trained model provided by Text2LIVE~\cite{text2live}.
    }
    \label{fig:comparison_text2live}
    % \vspace{-0.5cm} 
\end{figure}

%% file: Tables/quantity.tex
\begin{table}
  \centering
  \tabcolsep=0.2em
  \caption{Quantitative evaluation and user study \uppercase\expandafter{\romannumeral1} for baselines. The best results are highlighted in \textbf{bold} while the second best results are marked with an \underline{underline}.}
    \begin{tabular}{p{4.7em}|ccccc}
    \toprule
    \multicolumn{1}{c|}{} & Ours        & \multicolumn{1}{p{3.3em}}{Tune-A-Video} & Deforum     & Text2LIVE   & VideoCrafter \\
    \midrule
    \midrule
    Fra-Con↑ & \textbf{0.987} & 0.882       & 0.908       & 0.969      & \underline{0.973} \\
    \midrule
    Pro-Con↑ & \textbf{0.304} & 0.235       & 0.263       & \underline{0.272}       & 0.266 \\
    \midrule
    Fra-Acc↑ & \underline{0.983}       & 0.75        & 0.872       & \textbf{0.987} & 0.945 \\
    \midrule
    Preference↑ & - & 0.157       & 0.086       & 0.229       & 0.286 \\
    \bottomrule
    \bottomrule
    \end{tabular}%
  \label{tab:quantity}%
\end{table}%

%% file: Figures/userstudy.tex
\begin{figure}
    \centering
    \includegraphics[width=8.5cm]{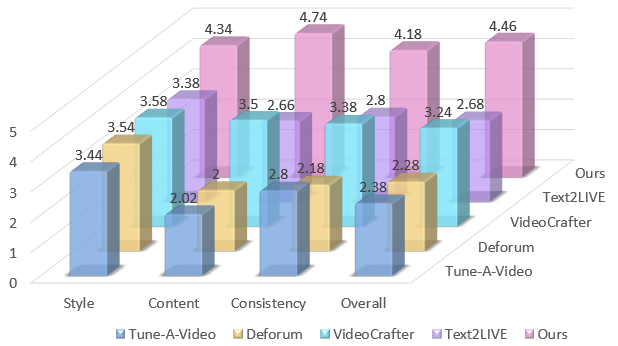}
    % \vspace{-0.2cm} 
    \caption{\textbf{User study \uppercase\expandafter{\romannumeral2}.} The average user rating results under the four metrics.}
    % \caption{Examples of failure cases are given source images, and textual guidance from ``coffee'' to ``water''.}
    % \vspace{-0.4cm} 
    \label{fig:userstudy}
\end{figure}

%% file: Figures/ablation_study.tex
\begin{figure}[tbp]
    \centering
    \includegraphics[width=8.5cm]{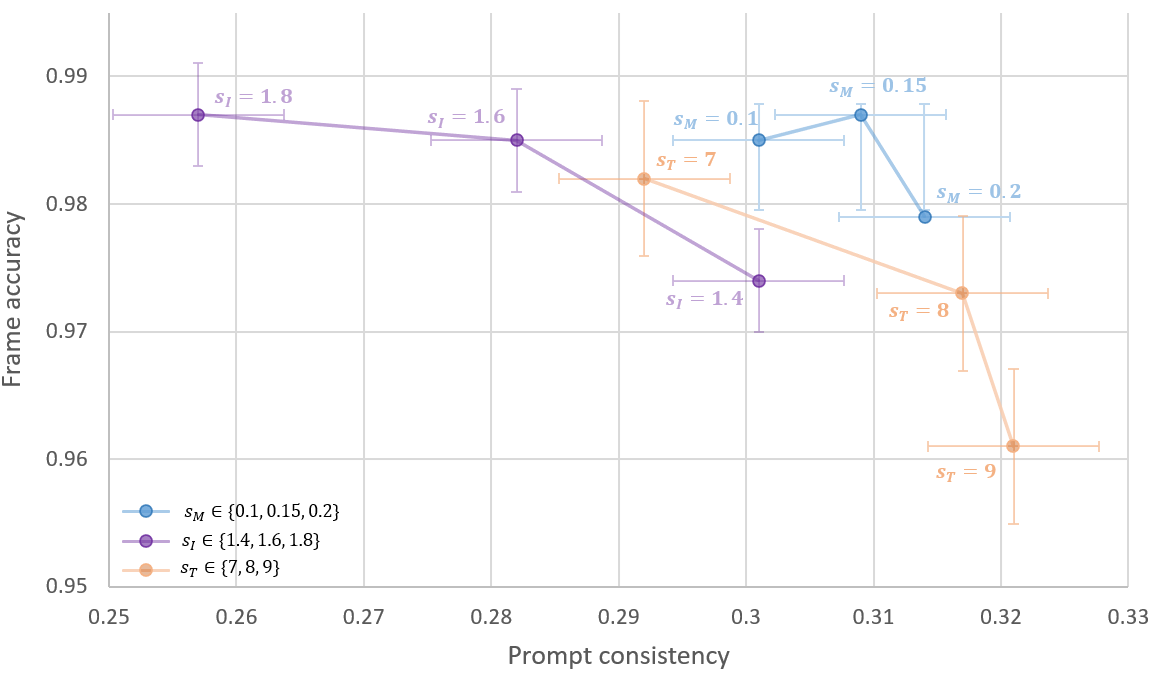}
    % \vspace{-0.2cm} 
    \caption{\textbf{Ablation study.} This figure reflects the effect of different condition guidance parameters scales on frame accuracy and prompt consistency. Better viewed for zoom-in.}
    % \vspace{-0.4cm} 
    \label{fig:ablation}
\end{figure}

%% file: Figures/ablation_wo.tex
\begin{figure}[tbp]
    \centering
    \includegraphics[width=8.5cm]{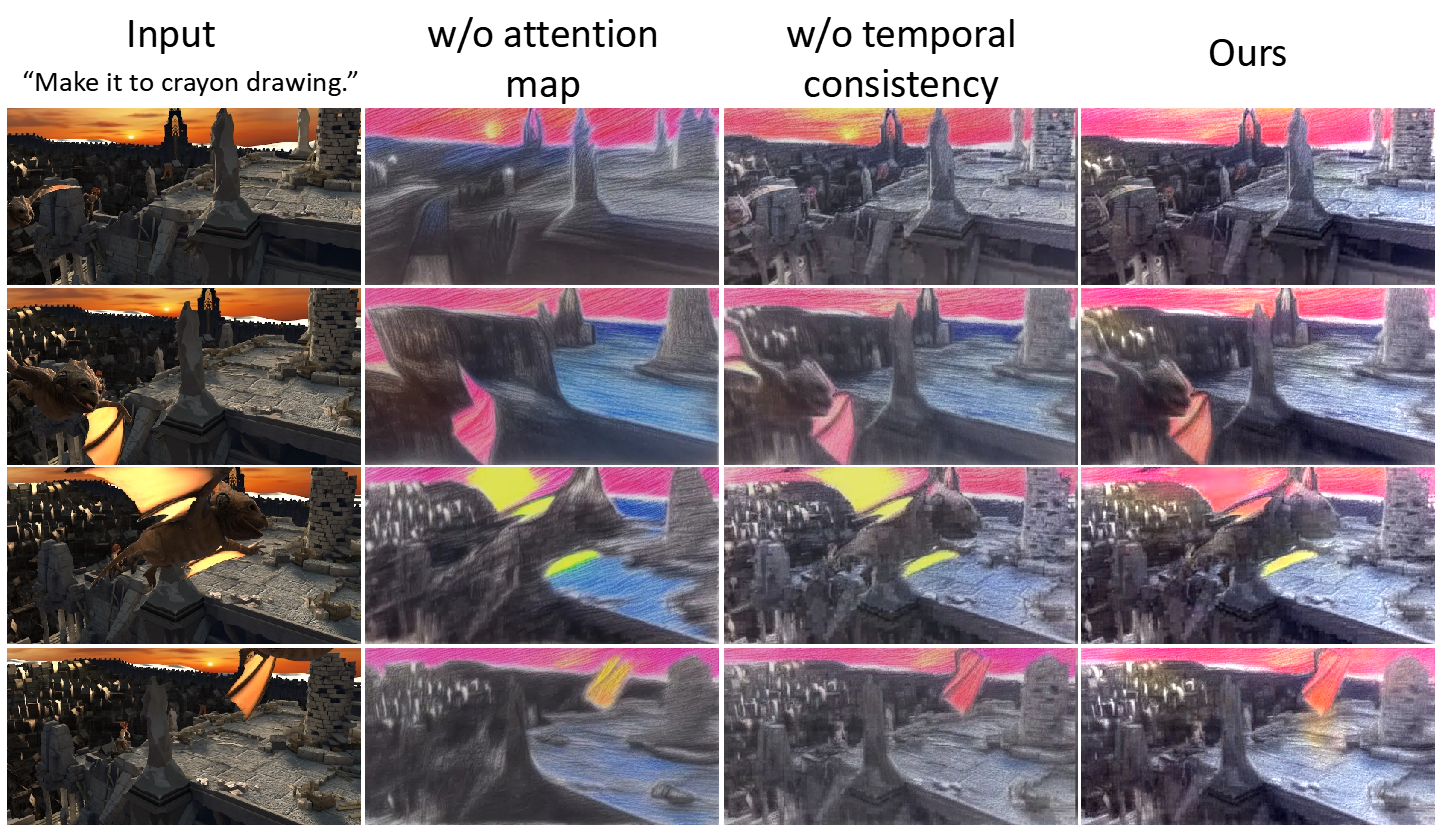}
    % \vspace{-0.2cm} 
    \caption{\textbf{Ablation study.} This figure presents the visualization of the impact of attention maps and temporal consistency on Style-A-Video.}
    % \caption{Examples of failure cases are given source images, and textual guidance from ``coffee'' to ``water''.}
    % \vspace{-0.4cm} 
    \label{fig:ablation_wo}
\end{figure}

%% file: Sections/6_conclusion.tex
\section{CONCLUSIONS AND FUTURE WORK}
In this paper, we propose a new text-driven video stylization framework Style-A-Video. 
It is based on the latent diffusion model and is capable of stylizing videos based on given style and content information. 
We implement to provide temporal consistency guidance, style guidance, and content retention guidance in each denoising process. 
We further propose the cross-attention module as a guiding condition for structural information to improve the performance of our framework in terms of structural retention and overall quality. 
We try to use text, video, and cross-attention graphs as conditions simultaneously for video stylization applications. 
Our framework is capable of generating arbitrary video, and arbitrary style, in a faster inference process. 
We believe this will pave the way for the wide application of video synthesis and editing. 
In the future, we plan to study the effect of other conditions such as three parameters and pose estimation on video stability.